\documentclass[sigconf]{acmart}
\AtBeginDocument{%
  }

\copyrightyear{2025}
\acmYear{2025}
\setcopyright{cc}
\setcctype{by}
\acmConference[WWW Companion '25]{Companion Proceedings of the ACM Web Conference 2025}{April 28-May 2, 2025}{Sydney, NSW, Australia}
\acmBooktitle{Companion Proceedings of the ACM Web Conference 2025 (WWW Companion '25), April 28-May 2, 2025, Sydney, NSW, Australia}
\acmDOI{10.1145/3701716.3717525}
\acmISBN{979-8-4007-1331-6/2025/04}

\usepackage{microtype}
\usepackage{multirow}
\usepackage{balance}

\begin{document}

\title{Identifying User Goals From UI Trajectories}

\author{Omri Berkovitch}
\authornote{Both authors contributed equally to this research.}
\orcid{0009-0009-4337-4954}
\affiliation{%
  \institution{Google Research}
    \city{Mountain View}
    \state{CA}
    \country{USA}
    }
\email{berkovitchomri@google.com}
\author{Sapir Caduri}
\orcid{0009-0009-2298-0605}
\affiliation{%
  \institution{Google Research}
    \city{Mountain View}
    \state{CA}
    \country{USA}
    }
\authornotemark[1]
\email{sapir@google.com}

\author{Noam Kahlon}
\orcid{0009-0004-0097-7417}
\affiliation{%
  \institution{Google Research}
    \city{Mountain View}
    \state{CA}
    \country{USA}
    }
\email{kahlonn@google.com}

\author{Anatoly Efros}
\orcid{0009-0000-1451-3438}
\affiliation{%
  \institution{Google Research}
    \city{Mountain View}
    \state{CA}
    \country{USA}
}
\email{talef@google.com}

\author{Avi Caciularu}
\orcid{0000-0003-0573-1075}
\affiliation{%
  \institution{Google Research}
    \city{Mountain View}
    \state{CA}
    \country{USA}
}
\email{avica@google.com}

\author{Ido Dagan}
\orcid{0009-0007-0165-606X}
\affiliation{%
  \institution{Google Research}
    \city{Mountain View}
    \state{CA}
    \country{USA}
}
\affiliation{%
  \institution{Bar-Ilan University}
    \city{Ramat Gan}
    \country{Israel}
}
\email{idodagan@google.com}






\renewcommand{\shortauthors}{Omri Berkovitch et al.}


\begin{abstract}
Identifying underlying user goals and intents has been recognized as valuable in various personalization-oriented settings, such as personalized agents, improved search responses, advertising, user analytics, and more. In this paper, we propose a new task—goal identification from observed UI trajectories—aiming to infer the user's detailed intentions when performing a task within UI environments. To support this task, we also introduce a novel evaluation methodology designed to assess whether two intent descriptions can be considered paraphrases within a specific UI environment. Furthermore, we demonstrate how this task can leverage datasets designed for the inverse problem of UI automation, utilizing Android and web datasets for our experiments. To benchmark this task, we compare the performance of humans and state-of-the-art models, specifically GPT-4 and Gemini-1.5 Pro, using our proposed metric. The results reveal that both Gemini and GPT underperform relative to human performance, underscoring the challenge of the proposed task and the significant room for improvement. This work highlights the importance of goal identification within UI trajectories, providing a foundation for further exploration and advancement in this area.


\end{abstract}



\received{20 February 2007}
\received[revised]{12 March 2009}
\received[accepted]{5 June 2009}

\maketitle





\section{Introduction}
\label{sec:introduction}
Autonomous agents that interact with GUIs to complete tasks for users have drawn increasing interest \cite{hong2023cogagent, gur2023real, yang2023appagent}. These agents interpret user-provided instructions and iteratively interact with GUIs to complete the desired task. In this work, we propose empowering agents with improved personalization abilities, allowing them to identify the underlying goals of users from their observed activity within the GUI environment. Understanding the underlying goals and intents of users is a cornerstone of user modeling, enabling agents to proactively align their behavior with user preferences, habits, and long-term objectives. This capability marks a significant step toward achieving personalization, where agents can provide context-aware, effective assistance that meets individual user needs \cite{li2024personal}.

\begin{figure}[t]
\centering
\includegraphics[trim=81pt 374.5pt 40pt 93pt, clip, width=1\columnwidth]{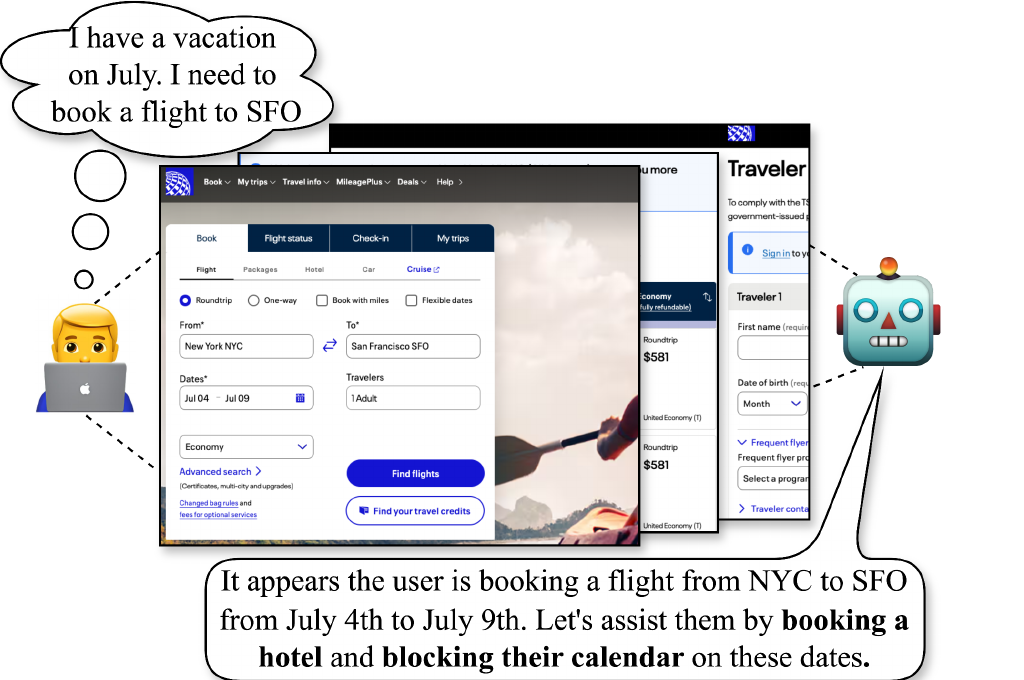}
\caption{An example of a user performing a flight booking task. The agent first observes the UI interactions, comprehends the task's essence, and then offers help with related tasks, like booking a hotel and blocking calendar dates. We focus on the first part, comprehending the task by observing the UI interactions.}
\Description{The figure illustrates a sequence of UI interactions in a flight booking scenario. A user selects flight details like destination and dates, and the agent interprets these actions. Additional UI interactions for related tasks, such as booking a hotel and calendar updates, are shown. The agent observes and understands these UI trajectories.}

\label{fig:task_illustration}
\end{figure}

Consider the scenario in Figure \ref{fig:task_illustration} where a user books flight tickets for a vacation. An ideal agent would observe these actions, understand the underlying user goal, and then proactively suggest booking a hotel for the same dates, and make the dates visible in the calendar. 

Our work extends a long line of research on recognizing user goals from their observed behaviour, including intent, activity, and plan recognition. However, most prior work addressed settings in which the input consists of natural language user utterances, like search queries, dialog utterances, or social media posts, while the task was perceived as structured classification, selecting a category label from a predefined list. Our work, to the best of our knowledge, is the first to identify user goals from UI interactions, while providing a natural language description of the user goal as output. We refer to Section \ref{app:related_work} for an extensive overview.

In this paper, we first define the task of generating a natural language description of underlying user goals from observed UI trajectories, that is, from multi-modal (text and screen image) traces of user-system interactions. A challenging aspect of the task is its inherent ambiguity, since multiple user goals can often lead to the same UI activity. Second, we observe that our task can be framed as the inverse of the known \textit{UI automation task}, where an agent needs to perform a sequence of UI actions given a natural language description of the user's goal \cite{li2023uinav, wen2023droidbot}. 
We further introduce manual and automatic evaluation protocols, assessing whether the predicted and gold task descriptions are paraphrases within the given UI context. 

Subsequently, we conducted experiments over both web and Android UI sessions, leveraging existing UI automation datasets while swapping the input and output roles. Over these data, we compared and analyzed the performance of humans and state-of-the-art multi-modal models, showing that there is substantial room for modeling improvements in future work. 

Overall, we offer the following contributions: (1) introducing and formalizing the task of goal identification from UI trajectories, highlighting its significance and potential applications; (2) demonstrating how existing datasets for UI automation can be adapted for this task by interpreting it as the "reversed" process of the other; (3) proposing both manual and automatic evaluation methodologies for performance assessment; and (4) evaluating the performance of both humans and state-of-the-art models on this task.

We propose that these contributions would trigger research on this timely challenge.




\section{Background and Related Work} \label{app:related_work}

In this section, we provide an overview of key related research areas that aim to learn personal user behaviors and preferences, specifically intent recognition, activity recognition, and plan recognition.

\paragraph{Intent Recognition} 
Intent recognition, also referred to as intent classification, is a well-established field focused on identifying user intentions based on specific inputs. Traditionally, most research in this domain has concentrated on textual inputs, such as individual messages or utterances, which are then categorized into predefined intent classes. For instance, studies like \cite{schuurmans2019intent, kuchlous2020short, larson2022survey} have worked on improving intent classification within dialogue systems across various domains, focusing on the classification of short texts or single utterances into predetermined intent categories (e.g. ``find a train from Barcelona to Madrid'' would be classified into a system intent called ''find\_train''). Others, classified social media posts to determine whether they express an intention to make a purchase \cite{gupta2014identifying, haque2019mining}. To enhance understanding of user queries and infer its underlying intent, some works leverage external knowledge sources to get better results \cite{hu2009understanding}.

In addition to text-based inputs, multimodal methods have been developed to incorporate various types of inputs. For example, \cite{zhang2021multimodal} and \cite{gonzaga2021multimodal} investigate the use of both images and text: the former analyzes social news content to identify marketing intents and to classify intent topics, while the latter combines image and textual data to classify social media posts, aiming to identify the writers' intent such as provocative, informative, promotive and more. \citet{zhang2022mintrec} introduces a multimodal dataset for intent recognition in TV series, where inputs include visual, auditory, and textual data, while the outputs are classified into one of 25 possible intent categories.

\paragraph{Activity Recognition}
Activity recognition, as defined by \citet{sukthankar2014plan} and closely related to intent recognition, involves identifying specific human activities based on a series of observations, often through sensor data. While intent recognition focuses on understanding user intentions, activity recognition mostly concerned with classifying physical actions. Typically, systems in this domain are developed to classify these activities using sensory inputs. \citet{khan2022human} uses a neural network trained on 2D skeletal data captured by a motion sensor to classify human poses; \citet{ijaz2022multimodal} integrated accelerometer signals along with skeletal data to recognize and categorize nurse activities into 12 distinct types. A prominent study \cite{kwapisz2011activity} developed a supervised learning algorithm that uses accelerometer signal from Android smartphones to classify physical activities like walking, jogging, sitting, and standing. Moreover, research such as \cite{liao2005location} utilizes GPS traces to classify human activity and label significant locations, employing relational Markov networks to achieve this.

\paragraph{Plan Recognition}
In a plan recognition problem, a system is given a series of actions performed by an agent and is expected to infer the overall plan or goal which explains those actions \cite{kautz1991formal}. The distinction between activity recognition and plan recognition is the difference between recognizing a single activity and recognizing the relationships between a set of such activities that result in a complete plan or goal. In this field, foundational works like \cite{charniak1993bayesian, goldman2013new} rely on structured inputs and a plan library — a collection of possible plans and their associated goals — to construct probabilistic models that infer the plan from observed actions. Alternative approaches, such as in \cite{hong2001goal}, develop algorithms that do not depend on plan libraries; instead, they utilizing in-domain knowledge and applying this to UNIX system commands to understand broader user goal. A hybrid approach is demonstrated \cite{granada2017hybrid}, where the authors combine CNN models to detect human activities from a video stream of meal preparation and apply symbolic reasoning with a plan library to recognize the overall plan (e.g., actions like "breaking eggs" and "mixing ingredients" might be recognized as part of a plan to create an omelet).

\section{Task Definition} \label{task_def}

Given an observed UI trajectory (see below) performed by a user with the intention to complete a certain task, our goal is to recover the user's original intent from the observable trajectory. As mentioned in \S~\ref{sec:introduction}, this setting is effectively the inverse problem of the known \textit{UI Automation} task. We therefore adopt their input and output definitions, swapping their roles, which enables the use of UI automation datasets for our task as well.

Accordingly, our input is a UI trajectory --- a sequence of individual interaction steps between the user and the system along the session. Each step consists of a snapshot of the UI content at that moment, along with the corresponding action the user took at that step (see Appendix \ref{appendix:datasets_and_models} for the specific trajectory formats). From this trajectory, our goal is to generate a natural language description that accurately captures the user's intended task. Within the scope of this paper, we address the core setting of the intent identification task and therefore assume that the observed UI trajectory indeed successfully fulfills the underlying user intent.

The intent identification task, similar to other text generation tasks like summarization where multiple valid outputs can exist, is inherently ambiguous, mostly because the same trajectory may fulfill multiple intents, often due to varying specificity levels of the original user intent (as captured in the dataset). For example, when observing a Sushi restaurant booking, the user might have asked for that specific Sushi place, or more broadly for \textit{some} restaurant in that area, or of that cuisine type. When a model identifies an intent from the given trajectory, we expect it to predict the most likely one, as reflected in the dataset distribution.

\section{Evaluation Methodology}
This section outlines our evaluation methodology. Given an input UI trajectory and a corresponding gold task description, we assess whether a \textit{predicted} task description matches the gold reference. We start with necessary definitions (Section \ref{sec:definitions}), followed by a human and automatic evaluation protocols (Sections \ref{sec:human_evaluation_protocol} and \ref{sec:automatic_eval}).

\subsection{Definitions}
\label{sec:definitions}

\paragraph{Task Fulfillment by a Trajectory} 
As mentioned in \S~\ref{task_def}, we assume an observed UI trajectory fulfills the underlying user's intended tasks. Therefore, a predicted task description not fulfilled by the input trajectory is erroneous and does not match the gold task description.

Inspired by the taxonomy proposed by \citet{zhou2023webarena}, we differentiate between fulfillment for information-seeking intents and transactional intents (e.g. purchasing an item or changing settings). For the latter, fulfillment is achieved upon successfully completing the specific requirement outlined in the task.
For information-seeking intents, fulfillment is achieved when the trajectory provides the necessary information sought in the user intent. We note that a fulfilling trajectory may provide some extra information beyond the intent, if such additional information is inherently bundled in the UI environment together with the sought information. 

\paragraph{Satisfaction Relation between Tasks} 
We next aim to specify a matching criterion between two task descriptions, specifically a predicted one and the corresponding gold-reference.
Given two task descriptions A and B and a UI environment, we say that A \textit{satisfies} B in that environment if every reasonable trajectory that fulfills A would also fulfill B. In essence, this means that completing task A necessarily results in completing task B, making B a more general task than A in that UI environment. For instance, the task ``Purchase the \underline{earliest} train ticket to Edinburgh'' satisfies the task ``Purchase a train ticket to Edinburgh'' but not vice versa.

Building on these definitions, we consider a predicted task description to successfully \textit{match} the gold description if the two mutually satisfy each other, and \textit{partial match} when only one satisfies the other. Essentially, matching tasks can be considered as paraphrases of the same intent within the context of the UI environment. For instance, the tasks ``Find a large dining table'' and ``Find a dining table for 10-12 people'' match each other if the UI environment considers 10-12 people as large.
We highlight the relation between \textit{match} and ambiguous trajectory, which fulfills multiple task descriptions that do not match each other.

\subsection{Human Evaluation Protocol}
\label{sec:human_evaluation_protocol}
As with many text generation tasks, human evaluation is essential due to the limited reliability of automatic evaluation metrics. In our case, the annotator observes the gold and predicted task descriptions, with the corresponding trajectory, and assesses (1) whether the trajectory fulfills the predicted task, and (2) whether each of the predicted and gold task descriptions satisfies the other. Additionally, annotators validated the given data instance, excluding from the evaluation noisy instances in the original datasets (details in Appendix \ref{appendix:experiments}).


To assess the quality of our proposed metric, we randomly sampled 50 instances from each of our datasets (see \S~\ref{dataset_section}) and generated task descriptions using our two baseline models (see \S~\ref{models_section}). We then measured the pairwise inter-annotator agreement among three of the authors, resulting in an average Cohen's Kappa of 0.79 and 0.77 for fulfillment and satisfaction judgments, respectively, in the web dataset, and 0.91 and 0.86 in the Android dataset. These agreement levels are considered high according to Kappa values, which justified our decision to manually evaluate each baseline model in our experiments using a single annotator.

\subsection{Automatic Evaluation Metric}
\label{sec:automatic_eval}

We propose utilizing a Large Multimodal Model (LMM) as an automatic evaluator for the satisfaction criteria. Recent advancements in LMMs, such as those demonstrated in \cite{he2024webvoyager} and \cite{pan2024autonomous}, have shown promising results in employing GPT-4 \cite{achiam2023gpt} to assess task completion by autonomous agents. Building on this, we leverage the latest GPT-4o model as the automatic evaluator, to determine whether two task descriptions are mutually satisfied in the context of the trajectory (prompt details are in Appendix \ref{app:instructions_and_prompts}).
Measuring agreement with our human evaluation yielded a Kappa value of 0.48 (moderate agreement), suggesting a potential utility of model-based automatic evaluation for development cycles while highlighting the need for manual evaluation.

\begin{table}[b!]
\caption{Manual evaluation of human-generated task descriptions, at the different levels of match criteria (determined by the satisfaction relation) between annotators (H$_{i}$) and gold task descriptions (G), as well as among annotators. (H) represents average scores.}
\label{table:human_performance_results}
\centering
\small
\setlength{\tabcolsep}{5pt} 
\renewcommand{\arraystretch}{1.2} 
\begin{tabular}{lcccc}
\specialrule{1.5pt}{0pt}{1pt} 
Dataset &  & Non Match & Partial Match & Match (↑) \\ 
\midrule
\multirow{2}{*}{Mind2Web} 
 & H vs. G & 0.07 & 0.13 & 0.80 \\ 
 & \hspace{2.75mm}H$_{1}$ vs. H$_{2}$ & 0.09 & 0.10 & 0.81 \\ 
\midrule
\multirow{2}{*}{AitZ} 
 & H vs. G & 0.02 & 0.22 & 0.76 \\ 
 & \hspace{2.75mm}H$_{1}$ vs. H$_{2}$ & 0.02 & 0.04 & 0.94 \\ 
\specialrule{1.5pt}{1pt}{0pt} 
\end{tabular}
\end{table}

\section{Datasets and Baseline Models}\label{sec:dataset_and_models_sec}


\begin{table*}[t]
\caption{Manual and automatic evaluation scores of fulfillment relation and the different levels of match criteria (determined by the satisfaction relation), between model predictions and gold task descriptions.}
\label{table:manual_and_automatic_results}
\centering
\footnotesize
\renewcommand{\arraystretch}{1.2} 
\setlength{\tabcolsep}{5pt} 
\begin{tabular}{llccccccc}
\specialrule{1.5pt}{0pt}{1pt} 
 &  & \multicolumn{4}{c}{\textbf{Manual Eval}} & \multicolumn{3}{c}{\textbf{Automatic Eval}} \\ 
\cmidrule(lr){3-6} \cmidrule(lr){7-9}
Dataset & Model & Fulfillment & Non Match & Partial Match & Match (↑) & Non Match & Partial Match & Match (↑) \\ 
\midrule
Mind2Web & GPT-4       & 0.86 & 0.14 & 0.42 & 0.44       & 0.18 & 0.40 & 0.42 \\
         & Gemini-1.5  & 0.87 & 0.20 & 0.22 & \textbf{0.58} & 0.18 & 0.28 & \textbf{0.54} \\ 
\midrule
AitZ     & GPT-4       & 0.78 & 0.22 & 0.19 & \textbf{0.59} & 0.32 & 0.34 & 0.34 \\
         & Gemini-1.5  & 0.88 & 0.17 & 0.26 & 0.57       & 0.37 & 0.27 & \textbf{0.36} \\ 
\specialrule{1.5pt}{1pt}{0pt} 
\end{tabular}
\end{table*}

\subsection{Evaluated Datasets}\label{dataset_section}


A suitable dataset for our task should contain natural language descriptions of user intents, along with traces of UI sessions that fulfill them. Aiming to satisfy this requirement, leveraging datasets that were created originally for UI automation is a natural choice. In these datasets, each instance was created by first specifying a particular user intent, followed by performing and recording a UI session that fulfills the given intent. Thus, a suitable dataset for our task of intent recognition is obtained from a UI automation dataset by swapping the input and output roles.

To that end, we explored several existing UI automation datasets but found that many of them do not fully align with the requirements of generic real-world intent recognition. For example, WebGUM operates in a simplified GUI environment while 
WebShop \cite{furuta2023multimodal, yao2022webshop} is limited to the shopping domain. RUSS and WebArena \cite{xu2021grounding, zhou2023webarena}, while providing user intents or instructions, do not possess the needed UI traces. PixelHelp and OmniACT \cite{li2020mapping, kapoor2025omniact} provide only UI-level procedural instructions rather than natural high-level intents which are independent of a particular UI environment (asking to click a button, filling a text field etc.). Lastly, RICO \cite{deka2017rico} lacks annotations for user intents.

Considering these limitations, we turned to datasets that do satisfy our task's requirements, focusing on two UI environments: web and Android. For the web environment, we utilize the Mind2Web dataset \cite{deng2023mind2web}, the most widely used benchmark for autonomous web agents. The dataset was created by curating diverse tasks across popular websites, with annotators performing a series of actions to complete the goal. For Android, we used the prominent Android in the Wild (AitW) \cite{rawles2023android} dataset, while focusing on its quality-filtered subset Android in the Zoo (AitZ) \cite{zhang2024android} (details in Appendix \ref{appendix:datasets_and_models}).

\subsection{Models}\label{models_section}

Given the multimodal nature of UI trajectories, our models must be adept at interpreting both text and images. We selected two state-of-the-art LMMs, Gemini 1.5 Pro \cite{reid2024gemini} and GPT-4. These models are at the forefront of handling combined text and image inputs and offer a sufficiently large context window for our experiments.

In our experiments, the model was guided through a Chain-of-Thought \cite{wei2022chain} process, to analyze the trajectory in a step-by-step manner (prompt in Appendix \ref{app:instructions_and_prompts}).

In web, inspired by SeeAct \cite{zheng2024gpt} we drew a red bounding box around the element that the user interacted with to guide the model's attention, as well as to break ambiguity where textual action descriptions were not sufficient. To further focus models, we also truncated lengthy web images based on the bounding box position. (details in Appendix \ref{appendix:experiments}). For Android, no special modifications needed, actions are overlaid on the screenshots, and the screenshots are naturally smaller.


\section{Experiments}
\label{sec:experiments}
\subsection{Human Performance Evaluation}
To asses task difficulty and establish a baseline for models, two NLP practitioners, unfamiliar with the datasets, independently composed 50 task descriptions from Mind2Web, and 50 from AitZ. An annotator then evaluated\footnote{During the evaluation it was observed that human tasks were inherently fulfilled by the trajectory.} them as explained in \S~\ref{sec:human_evaluation_protocol}.

The results, summarized in Table \ref{table:human_performance_results}, reveal more matches between the human annotators and the gold task descriptions in Mind2Web compared to AitZ. Upon analysis of the disagreements, we found that the gold descriptions in Android were often more specific than those provided by human annotators. In most cases, the gold tasks were already fulfilled by the trajectory, resulting in no clear interactions that indicate the user goal. For example, if the task was ``Turn WiFi on'' and the WiFi was already on, the annotators inferred a more general task, such as ``Show WiFi settings''. Further analysis of non-matching records across both datasets revealed that ambiguous trajectories caused human disagreements. Detailed analysis in Appendix \ref{appendix:err_analysis}.

\subsection{Model Performance Evaluation}
Model evaluation included manual assessment of 100 predictions from each dataset and automatic evaluation using the entire test sets: 1,013 from Mind2Web and 506 from AitZ (see Appendix \ref{appendix:experiments}).

The results, outlined in Table \ref{table:manual_and_automatic_results}, reveal that Gemini outperforms GPT on the Mind2Web dataset, achieving higher match scores but still falling short of humans. In AitZ, both models perform comparably. Gemini tends to be more specific and detailed than the gold references, while GPT often generates more general, abstract goals. Both models, in certain instances, misidentified the actual user intent (e.g. ``Watch top rated movie trailer'' vs. ``Watch The Dark Knight trailer'') and exhibit limitations in visual screen understanding, leading to missing details, incorporating irrelevant information and hallucinations of non-existent information. Detailed discussion in Appendix \ref{appendix:err_analysis}.

These results underscore the complexity of accurately capturing user intent. Our experiments also concluded few-shot learning. However, in our setting, this is challenging due to multiple images, textual inputs and the thought process within example. Adding a single example deteriorated results, likely because the large context size and the difference between the exemplar and the test trajectory.

\section{Conclusion and Future Work} \label{future_work}

In this paper, we introduced the novel task of identifying user goals from UI trajectories and proposed a reliable evaluation methodology. Our experiments on Android and web datasets reveal a significant gap between humans and state-of-the-art multi-modal models, which is analyzed in detail.

Future work may include fine-tuning experiments and enhancing visual understanding for UI environments. We propose testing our task by evaluating its utility on downstream tasks like agent personalization and suggestions. Additionally, expanding the scope to other GUIs like iOS and Windows would broaden the impact of our findings.

\section*{Limitations} \label{limitations}

Our study is subject to several limitations due to the nature of the datasets used. First, both datasets primarily include English-language websites accessed in the U.S., thereby limiting the study to English-language interactions only. Second, in real-world scenarios, we believe that (1) user trajectories may be interleaved between multiple tasks as users adjust their objectives in real time or are interrupted by other tasks, (2) users might have more ambiguous goals that evolve during their interaction with the user interface and (3) users might be less proficient with computers or phones, leading to noisier trajectories that are more challenging to identify and interpret intent from. Moreover, in the Mind2Web dataset, all tasks were limited to interactions within the same website, not encompassing multi-website tasks. This constraint may not fully represent the complexity of real-world web usage.

Lastly, this study focuses solely on Android and web environments. These environments might exhibit different task distributions compared to other user interface environments such as iOS and Windows, potentially limiting our findings for Android and web environments.

\section*{Ethical Considerations}

The development of autonomous agents, while holding great potential for innovation, raises important ethical considerations. Our research focuses on understanding user intent from recorded UI trajectories, and it is crucial to acknowledge the potential privacy implications of tracking user activity. Ensuring the security and protection of this sensitive data is of the utmost importance. Employing techniques like on-device processing, anonymization, or other privacy-preserving methods can help mitigate risks and ensure user data remains protected. It is essential for researchers and developers to proactively address these concerns to foster trust and responsible innovation in the field of autonomous agents.

\begin{acks}
We wish to thank Guy Rom, Xiang Deng, and Eliya Nachmani for their valuable feedback on our research. We also would like to express our gratitude to Noam Etzion-Rosenberg and Nir Shemy for their work in data annotation.
\end{acks}


\bibliographystyle{ACM-Reference-Format}
\balance
\bibliography{bibliography}


\begin{thebibliography}{42}


\ifx \showCODEN    \undefined \def \showCODEN     #1{\unskip}     \fi
\ifx \showDOI      \undefined \def \showDOI       #1{#1}\fi
\ifx \showISBNx    \undefined \def \showISBNx     #1{\unskip}     \fi
\ifx \showISBNxiii \undefined \def \showISBNxiii  #1{\unskip}     \fi
\ifx \showISSN     \undefined \def \showISSN      #1{\unskip}     \fi
\ifx \showLCCN     \undefined \def \showLCCN      #1{\unskip}     \fi
\ifx \shownote     \undefined \def \shownote      #1{#1}          \fi
\ifx \showarticletitle \undefined \def \showarticletitle #1{#1}   \fi
\ifx \showURL      \undefined \def \showURL       {\relax}        \fi
\providecommand\bibfield[2]{#2}
\providecommand\bibinfo[2]{#2}
\providecommand\natexlab[1]{#1}
\providecommand\showeprint[2][]{arXiv:#2}

\bibitem[Achiam et~al\mbox{.}(2023)]%
        {achiam2023gpt}
\bibfield{author}{\bibinfo{person}{Josh Achiam}, \bibinfo{person}{Steven Adler}, \bibinfo{person}{Sandhini Agarwal}, \bibinfo{person}{Lama Ahmad}, \bibinfo{person}{Ilge Akkaya}, \bibinfo{person}{Florencia~Leoni Aleman}, \bibinfo{person}{Diogo Almeida}, \bibinfo{person}{Janko Altenschmidt}, \bibinfo{person}{Sam Altman}, \bibinfo{person}{Shyamal Anadkat}, {et~al\mbox{.}}} \bibinfo{year}{2023}\natexlab{}.
\newblock \showarticletitle{Gpt-4 technical report}.
\newblock \bibinfo{journal}{\emph{arXiv preprint arXiv:2303.08774}} (\bibinfo{year}{2023}).
\newblock


\bibitem[Charniak and Goldman(1993)]%
        {charniak1993bayesian}
\bibfield{author}{\bibinfo{person}{Eugene Charniak} {and} \bibinfo{person}{Robert~P Goldman}.} \bibinfo{year}{1993}\natexlab{}.
\newblock \showarticletitle{A Bayesian model of plan recognition}.
\newblock \bibinfo{journal}{\emph{Artificial Intelligence}} \bibinfo{volume}{64}, \bibinfo{number}{1} (\bibinfo{year}{1993}), \bibinfo{pages}{53--79}.
\newblock


\bibitem[Deka et~al\mbox{.}(2017)]%
        {deka2017rico}
\bibfield{author}{\bibinfo{person}{Biplab Deka}, \bibinfo{person}{Zifeng Huang}, \bibinfo{person}{Chad Franzen}, \bibinfo{person}{Joshua Hibschman}, \bibinfo{person}{Daniel Afergan}, \bibinfo{person}{Yang Li}, \bibinfo{person}{Jeffrey Nichols}, {and} \bibinfo{person}{Ranjitha Kumar}.} \bibinfo{year}{2017}\natexlab{}.
\newblock \showarticletitle{Rico: A mobile app dataset for building data-driven design applications}. In \bibinfo{booktitle}{\emph{Proceedings of the 30th annual ACM symposium on user interface software and technology}}. \bibinfo{pages}{845--854}.
\newblock


\bibitem[Deng et~al\mbox{.}(2023)]%
        {deng2023mind2web}
\bibfield{author}{\bibinfo{person}{Xiang Deng}, \bibinfo{person}{Yu Gu}, \bibinfo{person}{Boyuan Zheng}, \bibinfo{person}{Shijie Chen}, \bibinfo{person}{Samuel Stevens}, \bibinfo{person}{Boshi Wang}, \bibinfo{person}{Huan Sun}, {and} \bibinfo{person}{Yu Su}.} \bibinfo{year}{2023}\natexlab{}.
\newblock \showarticletitle{Mind2Web: Towards a Generalist Agent for the Web}.
\newblock \bibinfo{journal}{\emph{arXiv preprint arXiv:2306.06070}} (\bibinfo{year}{2023}).
\newblock


\bibitem[Furuta et~al\mbox{.}(2023)]%
        {furuta2023multimodal}
\bibfield{author}{\bibinfo{person}{Hiroki Furuta}, \bibinfo{person}{Kuang-Huei Lee}, \bibinfo{person}{Ofir Nachum}, \bibinfo{person}{Yutaka Matsuo}, \bibinfo{person}{Aleksandra Faust}, \bibinfo{person}{Shixiang~Shane Gu}, {and} \bibinfo{person}{Izzeddin Gur}.} \bibinfo{year}{2023}\natexlab{}.
\newblock \showarticletitle{Multimodal web navigation with instruction-finetuned foundation models}.
\newblock \bibinfo{journal}{\emph{arXiv preprint arXiv:2305.11854}} (\bibinfo{year}{2023}).
\newblock


\bibitem[Goldman et~al\mbox{.}(2013)]%
        {goldman2013new}
\bibfield{author}{\bibinfo{person}{Robert~P Goldman}, \bibinfo{person}{Christopher~W Geib}, {and} \bibinfo{person}{Christopher~A Miller}.} \bibinfo{year}{2013}\natexlab{}.
\newblock \showarticletitle{A new model of plan recognition}.
\newblock \bibinfo{journal}{\emph{arXiv preprint arXiv:1301.6700}} (\bibinfo{year}{2013}).
\newblock


\bibitem[Gonzaga et~al\mbox{.}(2021)]%
        {gonzaga2021multimodal}
\bibfield{author}{\bibinfo{person}{Victor~Machado Gonzaga}, \bibinfo{person}{Nils Murrugarra-Llerena}, {and} \bibinfo{person}{Ricardo Marcacini}.} \bibinfo{year}{2021}\natexlab{}.
\newblock \showarticletitle{Multimodal intent classification with incomplete modalities using text embedding propagation}. In \bibinfo{booktitle}{\emph{Proceedings of the Brazilian Symposium on Multimedia and the Web}}. \bibinfo{pages}{217--220}.
\newblock


\bibitem[Granada et~al\mbox{.}(2017)]%
        {granada2017hybrid}
\bibfield{author}{\bibinfo{person}{Roger~L Granada}, \bibinfo{person}{Ramon~Fraga Pereira}, \bibinfo{person}{Juarez Monteiro}, \bibinfo{person}{Duncan Dubugras~Alcoba Ruiz}, \bibinfo{person}{Rodrigo~Coelho Barros}, {and} \bibinfo{person}{Felipe~Rech Meneguzzi}.} \bibinfo{year}{2017}\natexlab{}.
\newblock \showarticletitle{Hybrid activity and plan recognition for video streams}. In \bibinfo{booktitle}{\emph{Proceedings of the 31st. AAAI Conference: Plan, Activity and Intent Recognition Workshop, 2017, Estados Unidos.}}
\newblock


\bibitem[Gupta et~al\mbox{.}(2014)]%
        {gupta2014identifying}
\bibfield{author}{\bibinfo{person}{Vineet Gupta}, \bibinfo{person}{Devesh Varshney}, \bibinfo{person}{Harsh Jhamtani}, \bibinfo{person}{Deepam Kedia}, {and} \bibinfo{person}{Shweta Karwa}.} \bibinfo{year}{2014}\natexlab{}.
\newblock \showarticletitle{Identifying purchase intent from social posts}. In \bibinfo{booktitle}{\emph{Proceedings of the International AAAI Conference on Web and Social Media}}, Vol.~\bibinfo{volume}{8}. \bibinfo{pages}{180--186}.
\newblock


\bibitem[Gur et~al\mbox{.}(2023)]%
        {gur2023real}
\bibfield{author}{\bibinfo{person}{Izzeddin Gur}, \bibinfo{person}{Hiroki Furuta}, \bibinfo{person}{Austin Huang}, \bibinfo{person}{Mustafa Safdari}, \bibinfo{person}{Yutaka Matsuo}, \bibinfo{person}{Douglas Eck}, {and} \bibinfo{person}{Aleksandra Faust}.} \bibinfo{year}{2023}\natexlab{}.
\newblock \showarticletitle{A real-world webagent with planning, long context understanding, and program synthesis}.
\newblock \bibinfo{journal}{\emph{arXiv preprint arXiv:2307.12856}} (\bibinfo{year}{2023}).
\newblock


\bibitem[Haque et~al\mbox{.}(2019)]%
        {haque2019mining}
\bibfield{author}{\bibinfo{person}{Rejwanul Haque}, \bibinfo{person}{Arvind Ramadurai}, \bibinfo{person}{Mohammed Hasanuzzaman}, {and} \bibinfo{person}{Andy Way}.} \bibinfo{year}{2019}\natexlab{}.
\newblock \showarticletitle{Mining purchase intent in twitter}.
\newblock \bibinfo{journal}{\emph{Computaci{\'o}n y Sistemas}} \bibinfo{volume}{23}, \bibinfo{number}{3} (\bibinfo{year}{2019}), \bibinfo{pages}{871--881}.
\newblock


\bibitem[He et~al\mbox{.}(2024)]%
        {he2024webvoyager}
\bibfield{author}{\bibinfo{person}{Hongliang He}, \bibinfo{person}{Wenlin Yao}, \bibinfo{person}{Kaixin Ma}, \bibinfo{person}{Wenhao Yu}, \bibinfo{person}{Yong Dai}, \bibinfo{person}{Hongming Zhang}, \bibinfo{person}{Zhenzhong Lan}, {and} \bibinfo{person}{Dong Yu}.} \bibinfo{year}{2024}\natexlab{}.
\newblock \showarticletitle{WebVoyager: Building an End-to-End Web Agent with Large Multimodal Models}.
\newblock \bibinfo{journal}{\emph{arXiv preprint arXiv:2401.13919}} (\bibinfo{year}{2024}).
\newblock


\bibitem[Hong(2001)]%
        {hong2001goal}
\bibfield{author}{\bibinfo{person}{Jun Hong}.} \bibinfo{year}{2001}\natexlab{}.
\newblock \showarticletitle{Goal recognition through goal graph analysis}.
\newblock \bibinfo{journal}{\emph{Journal of Artificial Intelligence Research}}  \bibinfo{volume}{15} (\bibinfo{year}{2001}), \bibinfo{pages}{1--30}.
\newblock


\bibitem[Hong et~al\mbox{.}(2023)]%
        {hong2023cogagent}
\bibfield{author}{\bibinfo{person}{Wenyi Hong}, \bibinfo{person}{Weihan Wang}, \bibinfo{person}{Qingsong Lv}, \bibinfo{person}{Jiazheng Xu}, \bibinfo{person}{Wenmeng Yu}, \bibinfo{person}{Junhui Ji}, \bibinfo{person}{Yan Wang}, \bibinfo{person}{Zihan Wang}, \bibinfo{person}{Yuxiao Dong}, \bibinfo{person}{Ming Ding}, {et~al\mbox{.}}} \bibinfo{year}{2023}\natexlab{}.
\newblock \showarticletitle{Cogagent: A visual language model for gui agents}.
\newblock \bibinfo{journal}{\emph{arXiv preprint arXiv:2312.08914}} (\bibinfo{year}{2023}).
\newblock


\bibitem[Hu et~al\mbox{.}(2009)]%
        {hu2009understanding}
\bibfield{author}{\bibinfo{person}{Jian Hu}, \bibinfo{person}{Gang Wang}, \bibinfo{person}{Fred Lochovsky}, \bibinfo{person}{Jian-tao Sun}, {and} \bibinfo{person}{Zheng Chen}.} \bibinfo{year}{2009}\natexlab{}.
\newblock \showarticletitle{Understanding user's query intent with wikipedia}. In \bibinfo{booktitle}{\emph{Proceedings of the 18th international conference on World wide web}}. \bibinfo{pages}{471--480}.
\newblock


\bibitem[Ijaz et~al\mbox{.}(2022)]%
        {ijaz2022multimodal}
\bibfield{author}{\bibinfo{person}{Momal Ijaz}, \bibinfo{person}{Renato Diaz}, {and} \bibinfo{person}{Chen Chen}.} \bibinfo{year}{2022}\natexlab{}.
\newblock \showarticletitle{Multimodal transformer for nursing activity recognition}. In \bibinfo{booktitle}{\emph{Proceedings of the IEEE/CVF conference on computer vision and pattern recognition}}. \bibinfo{pages}{2065--2074}.
\newblock


\bibitem[Kapoor et~al\mbox{.}(2025)]%
        {kapoor2025omniact}
\bibfield{author}{\bibinfo{person}{Raghav Kapoor}, \bibinfo{person}{Yash~Parag Butala}, \bibinfo{person}{Melisa Russak}, \bibinfo{person}{Jing~Yu Koh}, \bibinfo{person}{Kiran Kamble}, \bibinfo{person}{Waseem AlShikh}, {and} \bibinfo{person}{Ruslan Salakhutdinov}.} \bibinfo{year}{2025}\natexlab{}.
\newblock \showarticletitle{Omniact: A dataset and benchmark for enabling multimodal generalist autonomous agents for desktop and web}. In \bibinfo{booktitle}{\emph{European Conference on Computer Vision}}. Springer, \bibinfo{pages}{161--178}.
\newblock


\bibitem[Kautz et~al\mbox{.}(1991)]%
        {kautz1991formal}
\bibfield{author}{\bibinfo{person}{Henry~A Kautz} {et~al\mbox{.}}} \bibinfo{year}{1991}\natexlab{}.
\newblock \showarticletitle{A formal theory of plan recognition and its implementation}.
\newblock \bibinfo{journal}{\emph{Reasoning about plans}} (\bibinfo{year}{1991}), \bibinfo{pages}{69--125}.
\newblock


\bibitem[Khan et~al\mbox{.}(2022)]%
        {khan2022human}
\bibfield{author}{\bibinfo{person}{Imran~Ullah Khan}, \bibinfo{person}{Sitara Afzal}, {and} \bibinfo{person}{Jong~Weon Lee}.} \bibinfo{year}{2022}\natexlab{}.
\newblock \showarticletitle{Human activity recognition via hybrid deep learning based model}.
\newblock \bibinfo{journal}{\emph{Sensors}} \bibinfo{volume}{22}, \bibinfo{number}{1} (\bibinfo{year}{2022}), \bibinfo{pages}{323}.
\newblock


\bibitem[Kuchlous and Kadaba(2020)]%
        {kuchlous2020short}
\bibfield{author}{\bibinfo{person}{Sahil Kuchlous} {and} \bibinfo{person}{Madhura Kadaba}.} \bibinfo{year}{2020}\natexlab{}.
\newblock \showarticletitle{Short text intent classification for conversational agents}. In \bibinfo{booktitle}{\emph{2020 IEEE 17th India Council International Conference (INDICON)}}. IEEE, \bibinfo{pages}{1--4}.
\newblock


\bibitem[Kwapisz et~al\mbox{.}(2011)]%
        {kwapisz2011activity}
\bibfield{author}{\bibinfo{person}{Jennifer~R Kwapisz}, \bibinfo{person}{Gary~M Weiss}, {and} \bibinfo{person}{Samuel~A Moore}.} \bibinfo{year}{2011}\natexlab{}.
\newblock \showarticletitle{Activity recognition using cell phone accelerometers}.
\newblock \bibinfo{journal}{\emph{ACM SigKDD Explorations Newsletter}} \bibinfo{volume}{12}, \bibinfo{number}{2} (\bibinfo{year}{2011}), \bibinfo{pages}{74--82}.
\newblock


\bibitem[Larson and Leach(2022)]%
        {larson2022survey}
\bibfield{author}{\bibinfo{person}{Stefan Larson} {and} \bibinfo{person}{Kevin Leach}.} \bibinfo{year}{2022}\natexlab{}.
\newblock \showarticletitle{A survey of intent classification and slot-filling datasets for task-oriented dialog}.
\newblock \bibinfo{journal}{\emph{arXiv preprint arXiv:2207.13211}} (\bibinfo{year}{2022}).
\newblock


\bibitem[Li et~al\mbox{.}(2023)]%
        {li2023uinav}
\bibfield{author}{\bibinfo{person}{Wei Li}, \bibinfo{person}{Fu-Lin Hsu}, \bibinfo{person}{Will Bishop}, \bibinfo{person}{Folawiyo Campbell-Ajala}, \bibinfo{person}{Oriana Riva}, {and} \bibinfo{person}{Max Lin}.} \bibinfo{year}{2023}\natexlab{}.
\newblock \showarticletitle{UINav: A maker of UI automation agents}.
\newblock \bibinfo{journal}{\emph{arXiv preprint arXiv:2312.10170}} (\bibinfo{year}{2023}).
\newblock


\bibitem[Li et~al\mbox{.}(2020)]%
        {li2020mapping}
\bibfield{author}{\bibinfo{person}{Yang Li}, \bibinfo{person}{Jiacong He}, \bibinfo{person}{Xin Zhou}, \bibinfo{person}{Yuan Zhang}, {and} \bibinfo{person}{Jason Baldridge}.} \bibinfo{year}{2020}\natexlab{}.
\newblock \showarticletitle{Mapping natural language instructions to mobile UI action sequences}.
\newblock \bibinfo{journal}{\emph{arXiv preprint arXiv:2005.03776}} (\bibinfo{year}{2020}).
\newblock


\bibitem[Li et~al\mbox{.}(2024)]%
        {li2024personal}
\bibfield{author}{\bibinfo{person}{Yuanchun Li}, \bibinfo{person}{Hao Wen}, \bibinfo{person}{Weijun Wang}, \bibinfo{person}{Xiangyu Li}, \bibinfo{person}{Yizhen Yuan}, \bibinfo{person}{Guohong Liu}, \bibinfo{person}{Jiacheng Liu}, \bibinfo{person}{Wenxing Xu}, \bibinfo{person}{Xiang Wang}, \bibinfo{person}{Yi Sun}, {et~al\mbox{.}}} \bibinfo{year}{2024}\natexlab{}.
\newblock \showarticletitle{Personal llm agents: Insights and survey about the capability, efficiency and security}.
\newblock \bibinfo{journal}{\emph{arXiv preprint arXiv:2401.05459}} (\bibinfo{year}{2024}).
\newblock


\bibitem[Liao et~al\mbox{.}(2005)]%
        {liao2005location}
\bibfield{author}{\bibinfo{person}{Lin Liao}, \bibinfo{person}{Dieter Fox}, {and} \bibinfo{person}{Henry Kautz}.} \bibinfo{year}{2005}\natexlab{}.
\newblock \showarticletitle{Location-based activity recognition}.
\newblock \bibinfo{journal}{\emph{Advances in neural information processing systems}}  \bibinfo{volume}{18} (\bibinfo{year}{2005}).
\newblock


\bibitem[Pan et~al\mbox{.}(2024)]%
        {pan2024autonomous}
\bibfield{author}{\bibinfo{person}{Jiayi Pan}, \bibinfo{person}{Yichi Zhang}, \bibinfo{person}{Nicholas Tomlin}, \bibinfo{person}{Yifei Zhou}, \bibinfo{person}{Sergey Levine}, {and} \bibinfo{person}{Alane Suhr}.} \bibinfo{year}{2024}\natexlab{}.
\newblock \showarticletitle{Autonomous evaluation and refinement of digital agents}.
\newblock \bibinfo{journal}{\emph{arXiv preprint arXiv:2404.06474}} (\bibinfo{year}{2024}).
\newblock


\bibitem[Rawles et~al\mbox{.}(2023)]%
        {rawles2023android}
\bibfield{author}{\bibinfo{person}{Christopher Rawles}, \bibinfo{person}{Alice Li}, \bibinfo{person}{Daniel Rodriguez}, \bibinfo{person}{Oriana Riva}, {and} \bibinfo{person}{Timothy Lillicrap}.} \bibinfo{year}{2023}\natexlab{}.
\newblock \showarticletitle{Android in the wild: A large-scale dataset for android device control}.
\newblock \bibinfo{journal}{\emph{arXiv preprint arXiv:2307.10088}} (\bibinfo{year}{2023}).
\newblock


\bibitem[Reid et~al\mbox{.}(2024)]%
        {reid2024gemini}
\bibfield{author}{\bibinfo{person}{Machel Reid}, \bibinfo{person}{Nikolay Savinov}, \bibinfo{person}{Denis Teplyashin}, \bibinfo{person}{Dmitry Lepikhin}, \bibinfo{person}{Timothy Lillicrap}, \bibinfo{person}{Jean-baptiste Alayrac}, \bibinfo{person}{Radu Soricut}, \bibinfo{person}{Angeliki Lazaridou}, \bibinfo{person}{Orhan Firat}, \bibinfo{person}{Julian Schrittwieser}, {et~al\mbox{.}}} \bibinfo{year}{2024}\natexlab{}.
\newblock \showarticletitle{Gemini 1.5: Unlocking multimodal understanding across millions of tokens of context}.
\newblock \bibinfo{journal}{\emph{arXiv preprint arXiv:2403.05530}} (\bibinfo{year}{2024}).
\newblock


\bibitem[Schuurmans and Frasincar(2019)]%
        {schuurmans2019intent}
\bibfield{author}{\bibinfo{person}{Jetze Schuurmans} {and} \bibinfo{person}{Flavius Frasincar}.} \bibinfo{year}{2019}\natexlab{}.
\newblock \showarticletitle{Intent classification for dialogue utterances}.
\newblock \bibinfo{journal}{\emph{IEEE Intelligent Systems}} \bibinfo{volume}{35}, \bibinfo{number}{1} (\bibinfo{year}{2019}), \bibinfo{pages}{82--88}.
\newblock


\bibitem[Sukthankar et~al\mbox{.}(2014)]%
        {sukthankar2014plan}
\bibfield{author}{\bibinfo{person}{Gita Sukthankar}, \bibinfo{person}{Christopher Geib}, \bibinfo{person}{Hung~Hai Bui}, \bibinfo{person}{David Pynadath}, {and} \bibinfo{person}{Robert~P Goldman}.} \bibinfo{year}{2014}\natexlab{}.
\newblock \bibinfo{booktitle}{\emph{Plan, activity, and intent recognition: Theory and practice}}.
\newblock \bibinfo{publisher}{Newnes}.
\newblock


\bibitem[Wei et~al\mbox{.}(2022)]%
        {wei2022chain}
\bibfield{author}{\bibinfo{person}{Jason Wei}, \bibinfo{person}{Xuezhi Wang}, \bibinfo{person}{Dale Schuurmans}, \bibinfo{person}{Maarten Bosma}, \bibinfo{person}{Fei Xia}, \bibinfo{person}{Ed Chi}, \bibinfo{person}{Quoc~V Le}, \bibinfo{person}{Denny Zhou}, {et~al\mbox{.}}} \bibinfo{year}{2022}\natexlab{}.
\newblock \showarticletitle{Chain-of-thought prompting elicits reasoning in large language models}.
\newblock \bibinfo{journal}{\emph{Advances in neural information processing systems}}  \bibinfo{volume}{35} (\bibinfo{year}{2022}), \bibinfo{pages}{24824--24837}.
\newblock


\bibitem[Wen et~al\mbox{.}(2023)]%
        {wen2023droidbot}
\bibfield{author}{\bibinfo{person}{Hao Wen}, \bibinfo{person}{Hongming Wang}, \bibinfo{person}{Jiaxuan Liu}, {and} \bibinfo{person}{Yuanchun Li}.} \bibinfo{year}{2023}\natexlab{}.
\newblock \showarticletitle{Droidbot-gpt: Gpt-powered ui automation for android}.
\newblock \bibinfo{journal}{\emph{arXiv preprint arXiv:2304.07061}} (\bibinfo{year}{2023}).
\newblock


\bibitem[Xu et~al\mbox{.}(2021)]%
        {xu2021grounding}
\bibfield{author}{\bibinfo{person}{Nancy Xu}, \bibinfo{person}{Sam Masling}, \bibinfo{person}{Michael Du}, \bibinfo{person}{Giovanni Campagna}, \bibinfo{person}{Larry Heck}, \bibinfo{person}{James Landay}, {and} \bibinfo{person}{Monica~S Lam}.} \bibinfo{year}{2021}\natexlab{}.
\newblock \showarticletitle{Grounding open-domain instructions to automate web support tasks}.
\newblock \bibinfo{journal}{\emph{arXiv preprint arXiv:2103.16057}} (\bibinfo{year}{2021}).
\newblock


\bibitem[Yang et~al\mbox{.}(2023b)]%
        {yang2023set}
\bibfield{author}{\bibinfo{person}{Jianwei Yang}, \bibinfo{person}{Hao Zhang}, \bibinfo{person}{Feng Li}, \bibinfo{person}{Xueyan Zou}, \bibinfo{person}{Chunyuan Li}, {and} \bibinfo{person}{Jianfeng Gao}.} \bibinfo{year}{2023}\natexlab{b}.
\newblock \showarticletitle{Set-of-mark prompting unleashes extraordinary visual grounding in gpt-4v}.
\newblock \bibinfo{journal}{\emph{arXiv preprint arXiv:2310.11441}} (\bibinfo{year}{2023}).
\newblock


\bibitem[Yang et~al\mbox{.}(2023a)]%
        {yang2023appagent}
\bibfield{author}{\bibinfo{person}{Zhao Yang}, \bibinfo{person}{Jiaxuan Liu}, \bibinfo{person}{Yucheng Han}, \bibinfo{person}{Xin Chen}, \bibinfo{person}{Zebiao Huang}, \bibinfo{person}{Bin Fu}, {and} \bibinfo{person}{Gang Yu}.} \bibinfo{year}{2023}\natexlab{a}.
\newblock \showarticletitle{Appagent: Multimodal agents as smartphone users}.
\newblock \bibinfo{journal}{\emph{arXiv preprint arXiv:2312.13771}} (\bibinfo{year}{2023}).
\newblock


\bibitem[Yao et~al\mbox{.}(2022)]%
        {yao2022webshop}
\bibfield{author}{\bibinfo{person}{Shunyu Yao}, \bibinfo{person}{Howard Chen}, \bibinfo{person}{John Yang}, {and} \bibinfo{person}{Karthik Narasimhan}.} \bibinfo{year}{2022}\natexlab{}.
\newblock \showarticletitle{Webshop: Towards scalable real-world web interaction with grounded language agents}.
\newblock \bibinfo{journal}{\emph{Advances in Neural Information Processing Systems}}  \bibinfo{volume}{35} (\bibinfo{year}{2022}), \bibinfo{pages}{20744--20757}.
\newblock


\bibitem[Zhang et~al\mbox{.}(2022)]%
        {zhang2022mintrec}
\bibfield{author}{\bibinfo{person}{Hanlei Zhang}, \bibinfo{person}{Hua Xu}, \bibinfo{person}{Xin Wang}, \bibinfo{person}{Qianrui Zhou}, \bibinfo{person}{Shaojie Zhao}, {and} \bibinfo{person}{Jiayan Teng}.} \bibinfo{year}{2022}\natexlab{}.
\newblock \showarticletitle{Mintrec: A new dataset for multimodal intent recognition}. In \bibinfo{booktitle}{\emph{Proceedings of the 30th ACM International Conference on Multimedia}}. \bibinfo{pages}{1688--1697}.
\newblock


\bibitem[Zhang et~al\mbox{.}(2024)]%
        {zhang2024android}
\bibfield{author}{\bibinfo{person}{Jiwen Zhang}, \bibinfo{person}{Jihao Wu}, \bibinfo{person}{Yihua Teng}, \bibinfo{person}{Minghui Liao}, \bibinfo{person}{Nuo Xu}, \bibinfo{person}{Xiao Xiao}, \bibinfo{person}{Zhongyu Wei}, {and} \bibinfo{person}{Duyu Tang}.} \bibinfo{year}{2024}\natexlab{}.
\newblock \showarticletitle{Android in the Zoo: Chain-of-Action-Thought for GUI Agents}.
\newblock \bibinfo{journal}{\emph{arXiv preprint arXiv:2403.02713}} (\bibinfo{year}{2024}).
\newblock


\bibitem[Zhang et~al\mbox{.}(2021)]%
        {zhang2021multimodal}
\bibfield{author}{\bibinfo{person}{Lu Zhang}, \bibinfo{person}{Jialie Shen}, \bibinfo{person}{Jian Zhang}, \bibinfo{person}{Jingsong Xu}, \bibinfo{person}{Zhibin Li}, \bibinfo{person}{Yazhou Yao}, {and} \bibinfo{person}{Litao Yu}.} \bibinfo{year}{2021}\natexlab{}.
\newblock \showarticletitle{Multimodal marketing intent analysis for effective targeted advertising}.
\newblock \bibinfo{journal}{\emph{IEEE Transactions on Multimedia}}  \bibinfo{volume}{24} (\bibinfo{year}{2021}), \bibinfo{pages}{1830--1843}.
\newblock


\bibitem[Zheng et~al\mbox{.}(2024)]%
        {zheng2024gpt}
\bibfield{author}{\bibinfo{person}{Boyuan Zheng}, \bibinfo{person}{Boyu Gou}, \bibinfo{person}{Jihyung Kil}, \bibinfo{person}{Huan Sun}, {and} \bibinfo{person}{Yu Su}.} \bibinfo{year}{2024}\natexlab{}.
\newblock \showarticletitle{Gpt-4v (ision) is a generalist web agent, if grounded}.
\newblock \bibinfo{journal}{\emph{arXiv preprint arXiv:2401.01614}} (\bibinfo{year}{2024}).
\newblock


\bibitem[Zhou et~al\mbox{.}(2023)]%
        {zhou2023webarena}
\bibfield{author}{\bibinfo{person}{Shuyan Zhou}, \bibinfo{person}{Frank~F Xu}, \bibinfo{person}{Hao Zhu}, \bibinfo{person}{Xuhui Zhou}, \bibinfo{person}{Robert Lo}, \bibinfo{person}{Abishek Sridhar}, \bibinfo{person}{Xianyi Cheng}, \bibinfo{person}{Yonatan Bisk}, \bibinfo{person}{Daniel Fried}, \bibinfo{person}{Uri Alon}, {et~al\mbox{.}}} \bibinfo{year}{2023}\natexlab{}.
\newblock \showarticletitle{Webarena: A realistic web environment for building autonomous agents}.
\newblock \bibinfo{journal}{\emph{arXiv preprint arXiv:2307.13854}} (\bibinfo{year}{2023}).
\newblock


\end{thebibliography}

\appendix

\section{Datasets Overview} \label{appendix:datasets_and_models}
\begin{figure*}[t]
\centering
\includegraphics[trim=15pt 100pt 30pt 160pt, clip, width=1\textwidth]{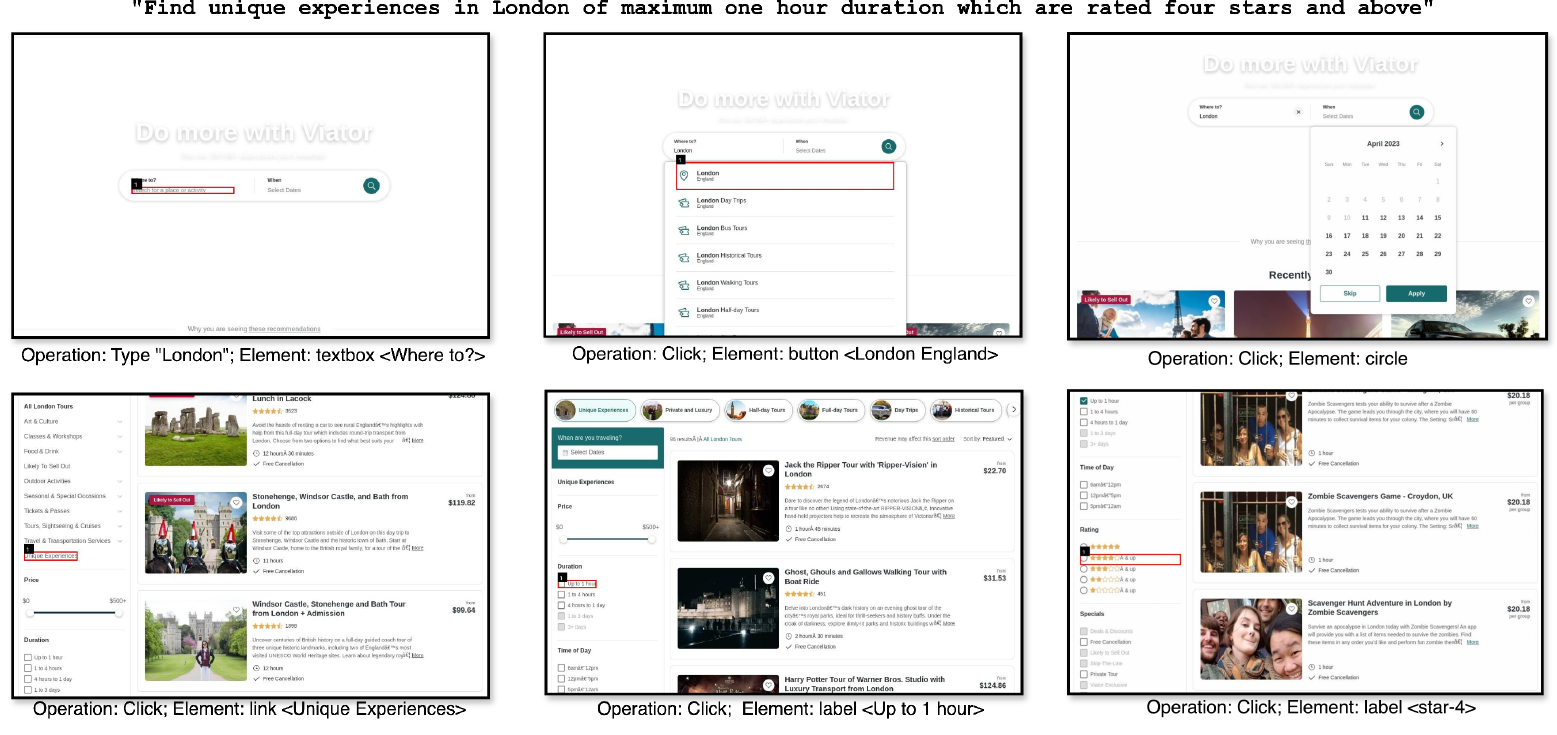}
\caption{An instance from Mind2Web, representing a full trajectory accomplishing the task description above.}
\Description{The figure depicts a sequence of user interactions within a web environment to complete the task of "Find unique experiences in London of maximum one hour duration which are rated four stars and above". It consists of six sub-images, each showcasing an web screen representing a specific step in the user's actions. This example is taken from the Mind2Web dataset.}
\label{fig:mind2web_example}
\end{figure*}

\begin{figure*}[t]
\centering
\includegraphics[scale=0.2]{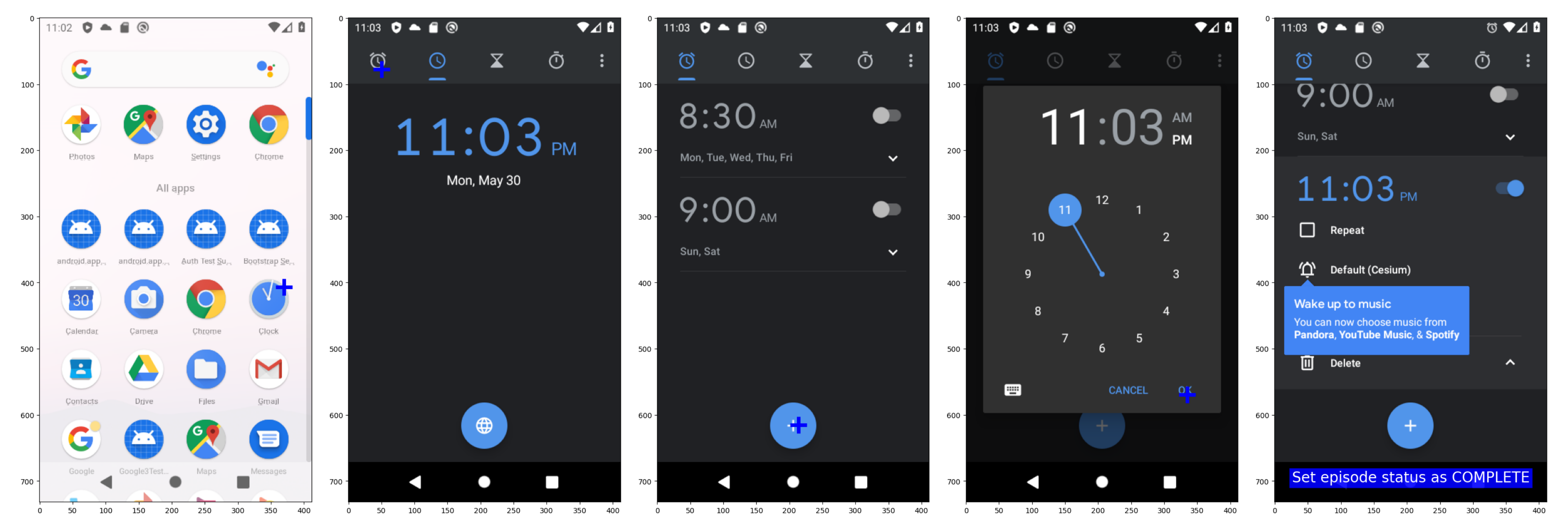} 

\caption{An instance from AitW, representing a full trajectory accomplishing the task "Set an alarm". The blue plus sign indicates the area on the screen where the tap occurred.}
\Description{The figure depicts a sequence of user interactions within an Android environment to complete the task of "Set an Alarm." It consists of five sub-images, each showcasing an Android screen representing a specific step in the user's actions. This example is taken from the Android In the Wild dataset.}
\label{fig:aitw_example}
\end{figure*}

\subsection{Web} We aim to give a brief overview of how Mind2Web was collected, and its format in more depth. The data collection process involved selecting numerous popular websites across five top-level domains. Annotators, guided by seed tasks from GPT, proposed diverse tasks for each website. They demonstrated and recorded how to complete these tasks, resulting in a trajectory of actions and screenshots. Each action is defined by a pair consisting of a Target Element and an Operation. The Target Element is an interactable element on the current web page, such as buttons, input fields, or drop down menus. The Operation refers to the specific action to be executed on the Target Element, with Mind2Web supporting three primary operations: Click (including actions like clicking, hovering, and pressing Enter), Type (which involves entering text into input fields and requires an additional value for the text to be typed), and Select Option (which involves selecting an option from a drop down menu or similar element and also requires an additional value for the option to be selected). Notably, these actions were automatically produced during the time users were recorded completing the task, eliminating the need for additional manual labor. Figure \ref{fig:mind2web_example} demonstrates a single instance of the data, where textual action descriptions are presented below the corresponding screenshot.

\subsection{Android}
The Android in the Wild (AitW) dataset stands out due to its extensive variety of tasks, covering 4 domains: Google-Apps, Install, Web-Shopping, and General, as well as a single-step domain excluded from this paper's analysis. The dataset consists of a substantial collection of high-level instructions, trajectories of varying lengths, and a notable variety of apps and websites. The Android in the Zoo (AitZ) dataset was sampled from AitW to reduce tasks redundancy and to filter erroneous data points, resulting in more unique and higher-quality task descriptions.

The high-level goal instructions in AitW were sourced from various sources, including humans (both crowd-sourced and the authors themselves), LLM-generated prompts, and technical documentation such as PixelHelp \cite{li2020mapping}. The creation process involved human annotators performing tasks on Android emulators, with their gestures being recorded.

The episodes were recorded on mobile devices running four different versions of Android. Each episode contains natural language instructions and observation-action pairs. The observations are screenshots, while the actions are one of three types: tap, drag, or typing. Gesture actions are represented as taps and drags at specific <x,y> coordinates on the screen.

In our utilization of this data, we presented the models with a series of screenshots with the actions drawn on them, as shown in Figure \ref{fig:aitw_example}.

\section{Experiments}\label{appendix:experiments}

\subsection{Models Configuration}
For task description prediction, we used the most recent version of Gemini, namely Gemini 1.5 Pro, updated in Vertex AI\footnote{https://cloud.google.com/vertex-ai} as of May 2024, with a sampling temperature of 1.0, and GPT-4-Turbo\footnote{https://platform.openai.com/docs/models/gpt-4-turbo-and-gpt-4} (version gpt-4-turbo-2024-04-09) with a sampling temperature of 0.6. For automatic evaluation, we utilized the latest release from the GPT family, known as GPT-4o.

\subsection{Data}
For the manual evaluation, we randomly sampled over 100 data points from the both Android and Web datasets. We then prompted models to predict the user goal. During the evaluation, we conducted a thorough verification process to ensure that both the gold references and the trajectories were of high quality.

We rejected instances where the trajectory did not fulfill the gold reference, which occurred more frequently for Android. Additionally, for web-based tasks, we excluded cases where the screenshot was not rendered properly, making it difficult to interpret the user’s action, as well as instances where the bounding box was empty.

As a result of this verification process, 17\% of the Android examples and about 5\% of the web examples were rejected. Ultimately, as a result of this process, the samples that were manually evaluated in Sections \ref{sec:human_evaluation_protocol} and \ref{sec:experiments} were of high quality and free from such issues. For the automatic evaluation, as described in \ref{sec:experiments}, we utilized the entire test sets from both datasets.

\subsection{Web} We encountered a technical challenge with the Mind2Web data due to the nature of its image captures. Unlike standard viewport captures, which represent the visible area of a web page on a typical screen, the images in Mind2Web had a median height of 4200 pixels, significantly exceeding typical web page dimensions, with 20 percent of the images exceeding 7000 pixels in height. Initial tests showed that these oversized images introduced noise and negatively affected model performance.

To address this, we implemented a heuristic truncation method, reducing image height while ensuring the interaction element remained visible within the truncated image. This was achieved by utilizing the bounding box metadata provided by Mind2Web. Similar to the approach taken by \cite{zheng2024gpt} and introduced by \cite{yang2023set}, we drew the red bounding box around the interaction element to guide the model's attention. Additionally, adding the bounding box helped resolve ambiguities at times that the textual action description is not sufficient. For example, sometimes action descriptions are simply empty, and does not contain any description about the element itself. While in other cases, action descriptions exist, but the textual description matches multiple element descriptions and thus results in ambiguity that is only resolved by drawing the bounding box. For example, a button labeled "Add to Cart" is typically associated with each item in a web shopping list. Without the bounding box, it is impossible to determine which specific button was clicked.

\subsection{Android} 
In our efforts to replicate experiments from web, we encountered challenges due to differences in data format. Unlike the web, AitW does not provide textual information associated with clicked elements. Instead, it offers x,y coordinates representing the center of the tapped area. This distinction in data structure made it impractical to conduct experiments involving both actions written alongside screenshots.

To address these challenges, we utilized the dataset's utilities to overlay actions on top of the screenshots, as well as bounding box annotations for post-process detected UI elements. However, we found that the added element annotation marks often confused the model added noise and critical information on the screen. As a result, we proceeded with experiments using a sequence of screenshots that had the actions (tap, drag, and type) drawn over them.
AitW's provided visualization tools to draw the actions, also included labels of special actions such as the back button, home button, and enter. As well as a special "status" action: either Task Complete or Task Impossible. However, we found that the label Task Complete often confused the models. Despite our efforts to make the models ignore it, we eventually abandoned this specific annotation as it is a technicality of data representation.
\section{Error Analysis} \label{appendix:err_analysis}
\begin{figure}[ht!]
\centering
\includegraphics[trim=50pt 315pt 30pt 80pt, clip, width=\columnwidth]{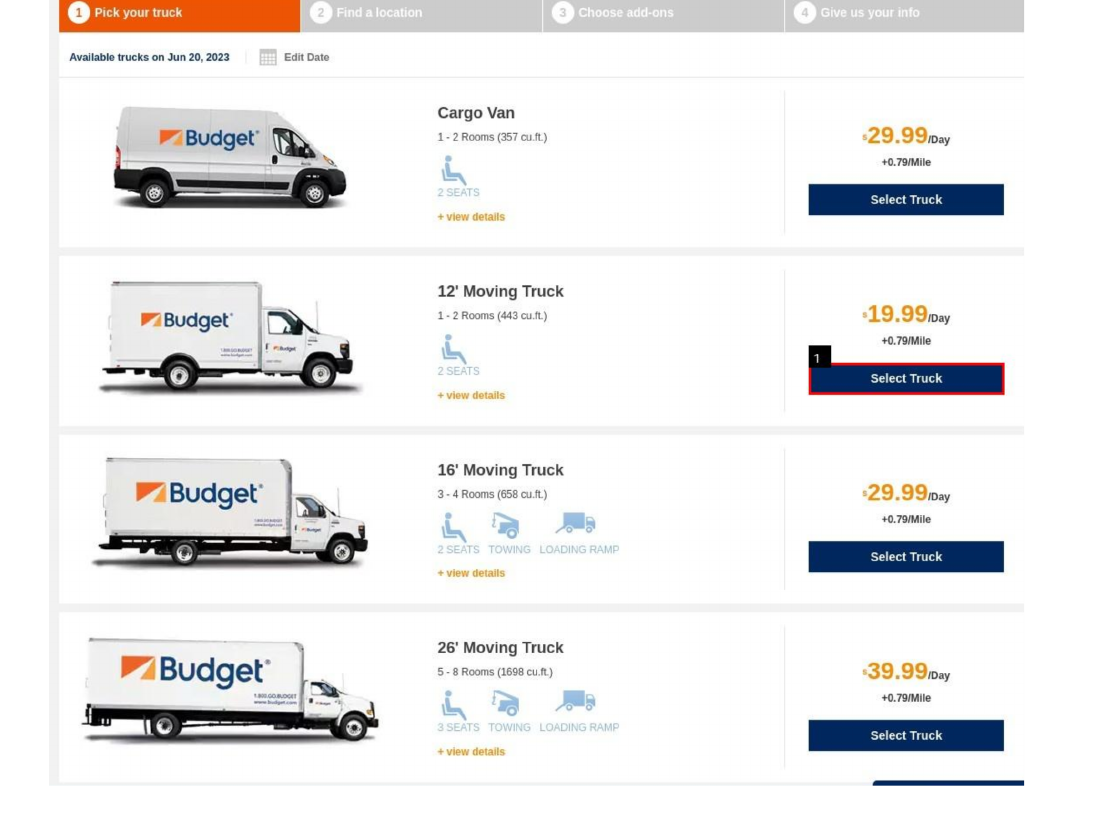}
\caption{An illustration in which the user chose a specific car primarily for its 12-inch feature, but since it was also the cheapest, annotators incorrectly assumed cost was the deciding factor.}
\Description{The figure presents an example from the Mind2Web dataset, showcasing four different trucks, each with distinct wheel sizes and prices. The image is annotated with the user's selection: the second truck, which has the lowest price and features 12'' wheels.}
    \label{fig:12_inch_truck_issue}
\end{figure}
In this section, we aim to provide a more detailed error analysis with respect to model and humans.

\subsection{Web} In our error analysis of 30 mismatched task descriptions, distinct patterns emerged between GPT and Gemini models. GPT frequently wrote tasks as navigational procedures, with 12\% of generated tasks starting with ``Navigate'' (out of the 1,000 predicted tasks). Additionally, over 20\% of the manually inspected errors involved misinterpreting the task's intent, often producing broad descriptions lacking crucial details. Conversely, Gemini's errors were typically more fine-grained, often capturing the task's essence but struggling with specific details like dates, numbers, or locations.

Both models also occasionally produced task descriptions that felt artificial, incorporating information a user would be unlikely to know beforehand due to the models' access to the full user trajectory. For example, a task like ``Read recent news about Apple stock'' might be predicted as ``Read the article 'X' about Apple stock'' if the model observed the user clicking on a specific article 'X'.

With respect to human annotators, we found that most disagreements between human-generated task descriptions and gold task descriptions resulted from humans making more generalized tasks. This happened because they choose the most natural or probable constraints if no action provides evidence for a less likely constraint. Sometimes, they don't write the constraint or any other one if nothing seems probable. Figure \ref{fig:12_inch_truck_issue} demonstrates such a case, the truck picked by the user is the only 12-inch wheel truck but also the cheapest truck among the listed options, as no prior action gave evident to the 12-inch constraint, both human generated task labeled it as ``Book the cheapest truck...'' while the gold task description was ``Book a 12-inch wheel truck...''. Additionally, with respect to task ambiguity, each trajectory in this experiment resulted in three task descriptions, two from the human annotators and one from the gold reference. We calculated the number of trajectories where all task descriptions matched each other and those where they did not. We found that 72\% of trajectories had a match among all tasks, 24\% had a match between two tasks, and the rest had no matching tasks. Although not exhaustive, this highlights that most trajectories in Mind2Web are probably non-ambiguous.

\begin{figure}[ht!]
\centering
\includegraphics[width=\columnwidth]{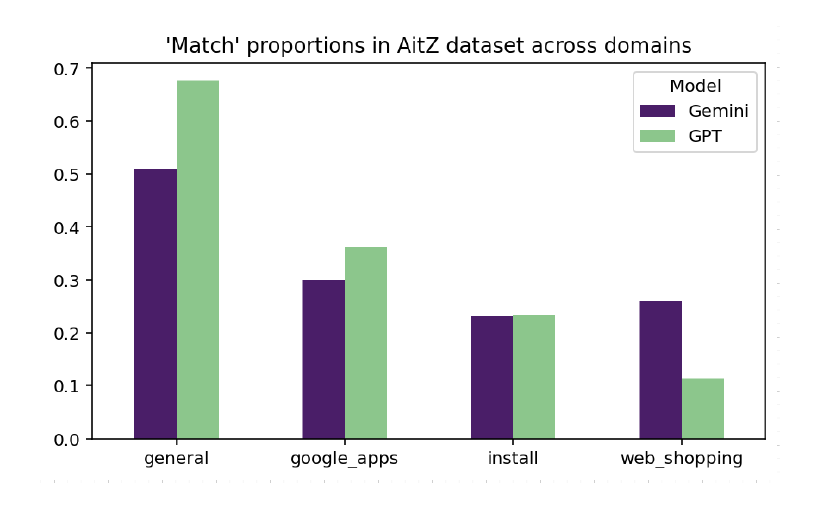}
\caption{Comparison of "Match" proportions between Gemini 1.5 Pro and GPT-4-Turbo models across the different domains}
\Description{The figure presents a histogram comparing the "Match" metric values across different domains in the Android in the Zoo dataset. The domains include "General," "Google Apps," "Install," and "Web Shopping." The comparison is made between the GPT-4 and Gemini 1.5 Pro models.}
    \label{fig:aitw_match_per_domain}
\end{figure}

\subsection{Android} 
Comparing GPT and Gemini's performance on the Android dataset revealed notable variations across different domains. Both models were proficient in the ``General'' domain, but faced challenges in ``Web Shopping'' and ``Install''. This disparity is due to the nature of the ``General'' dataset, which contains search queries that explicitly reveal the user's intent. ``Google Apps'' tasks overlap with PixelHelp, primarily involving settings configuration. Some tasks are ambiguous, such as a button toggle, but the trajectory displays only one option. Additionally, specific tasks request actions that have already been completed, e.g Turn On location history, but the trajectory only shows viewing the Location History setting page, leading to confusion for models.
Additionally, the models faced difficulty comprehending the correct order of the sequence of actions that occurred, frequently mistaking the final state (on or off).

On the ``Web Shopping'' domain, Gemini provided excessive details about specific products (``Add Razer Kraken X for Console Gaming Headset for PC/PS4/PS5/Xbox/Switch - Black/Blue to cart on Best Buy Canada'').
GPT often missed the main purpose of the task and suggested abstract tasks such as: ``log into an account'' or ``Decline the offer to protect a purchase with an insurance plan on a shopping website''.

The ``Install'' domain often presents ambiguous tasks in the format ``open (install if not installed)'' which confuses both models. Furthermore, in some cases the apps were already pre-installed which made it impossible to predict (model predicts ``open''), providing only partial satisfaction in one direction.  
These results indicate that further refinement and training may be needed to improve the models' performance in specific domains.

For human generated tasks, ``General'' dataset presented minimal challenges for human annotators. This was attributed to their ability to effortlessly comprehend the user's intended intent based solely on the visible search query. However, analogous to the model challenges encountered, ambiguous tasks within the ``Install'' dataset proved challenging for humans as well. Conversely, unlike models, humans exhibited impeccable performance in comprehending the final state of the desired setting configuration within the Google Apps domain, if the original goal was specific and not ambiguous. Shopping tasks, on the other hand, posed a distinct challenge for humans. They struggled to grasp the rationale behind selecting an item when the original task was to choose the cheapest or the first result. These findings underscore the multifaceted nature of goal task prediction and emphasize the significance of addressing specific domains.

\section{Instructions and Prompts} \label{app:instructions_and_prompts}

\begin{figure*}[ht!]
\centering
\includegraphics[trim=0pt 80pt 0pt 70pt, clip, width=\textwidth]{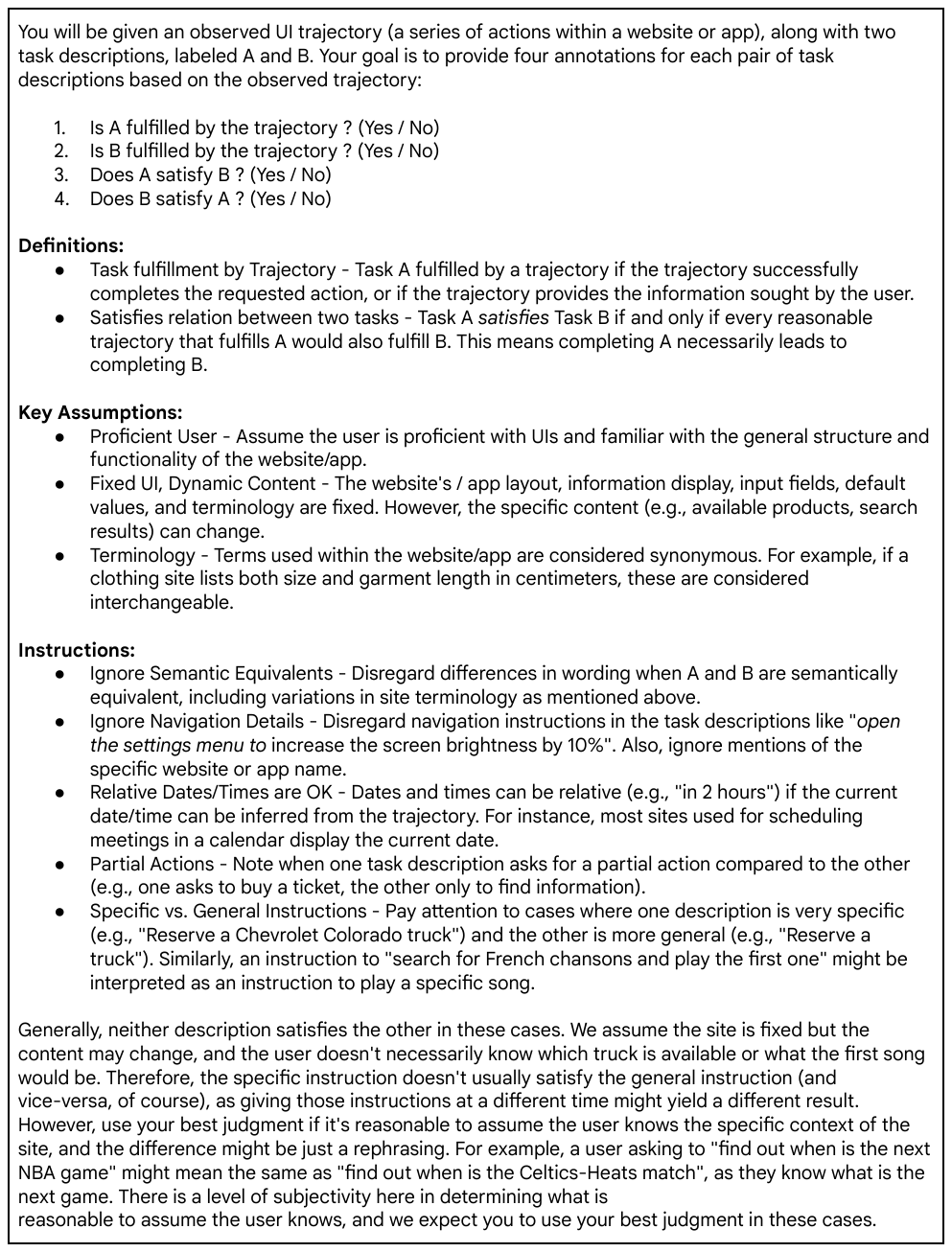}
\caption{Instructions for human-annotators to conclude if a task is fulfilled by a trajectory, and if two task descriptions satisfying each other. These instructions were the core prompt of the automatic evaluator.}
\Description{The figure contains text that was used to instruct human participants on how to evaluate the fulfillment and satisfaction relations when comparing two task descriptions with a UI trajectory.}
\label{fig:satisfies_fulfillment_instructions}
\end{figure*}

\begin{figure*}[ht]
\centering
\includegraphics[trim=50pt 230pt 30pt 105pt, clip, width=0.9\textwidth]{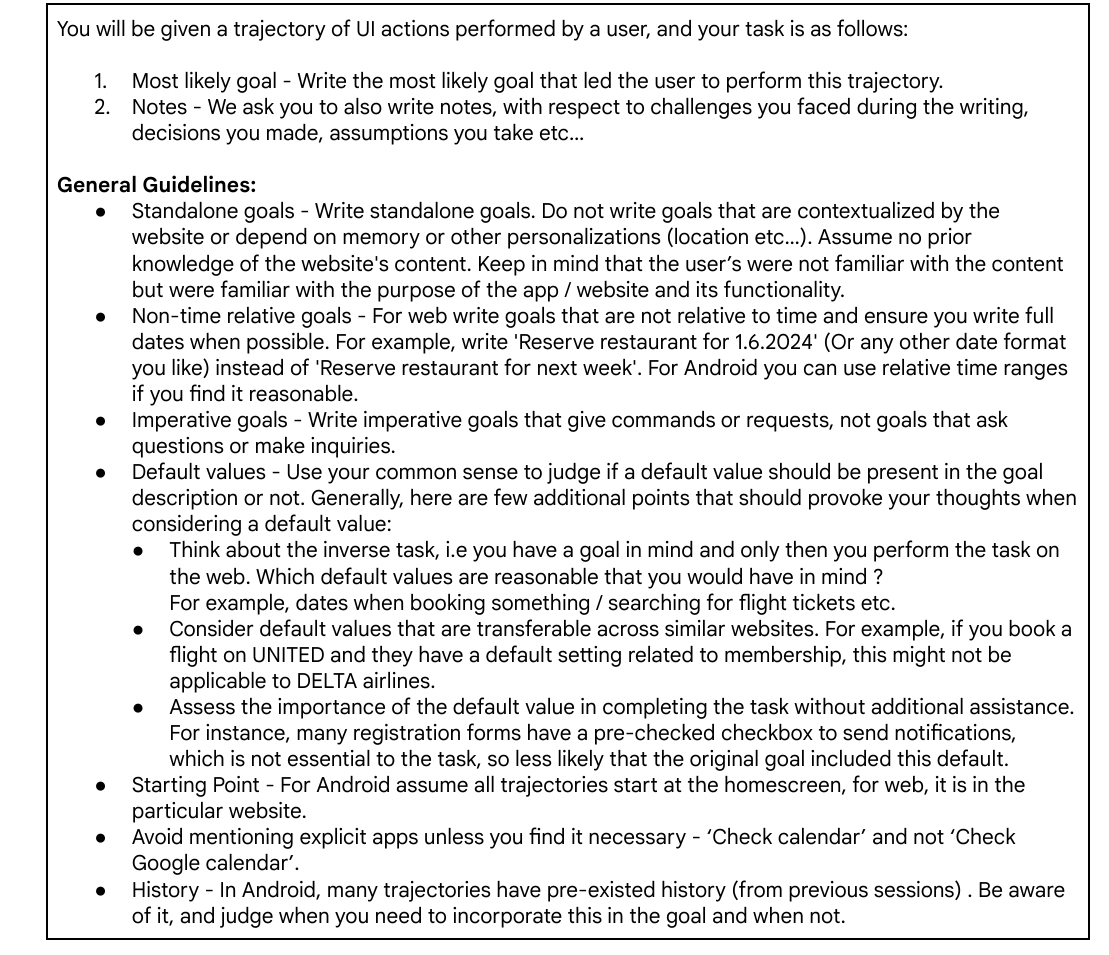}
\caption{Goal identification annotation task instruction to measure human performance.}
\Description{The figure contains text that was used to instruct human participants while performing the task of identifying user goals from UI trajectories.}
\label{fig:most_likely_instructions}
\end{figure*}

\begin{figure*}[ht]
\centering
\includegraphics[trim=0pt 130pt 0pt 70pt, clip, width=\textwidth]{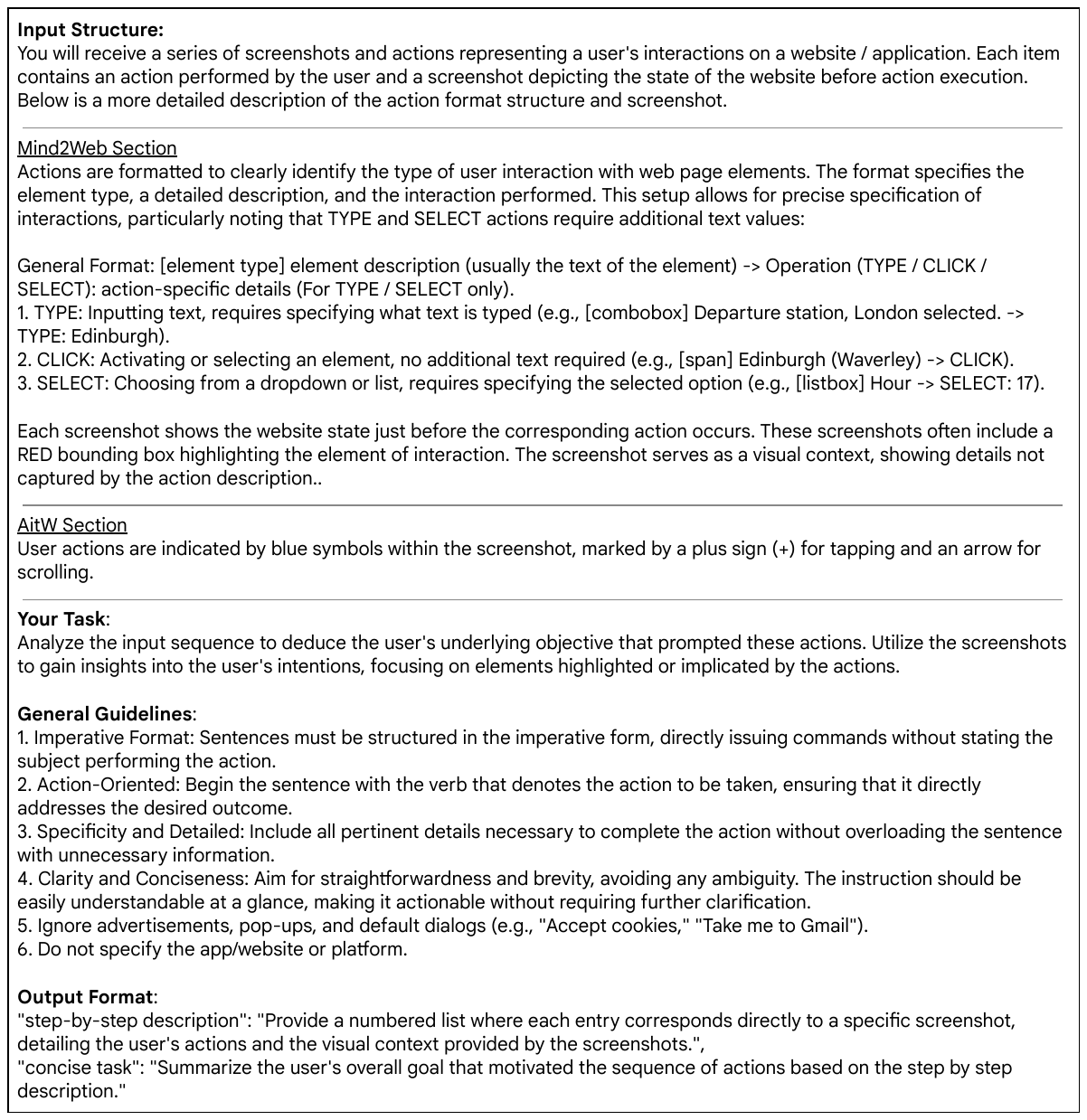}
\caption{Instruction used to guide GPT and Gemini when predicting a task description given a UI trajectory. We swap AitW section and Mind2Web according to the input.}
\Description{The figure contains text that was used to guide Gemini 1.5 Pro and GPT-4 models in predicting user goals based on UI trajectories.}
\label{fig:task_description_gen_prompt}
\end{figure*}

\begin{figure*}[t]
\centering
\includegraphics[trim=0pt 90pt 0pt 70pt, clip, width=\textwidth]{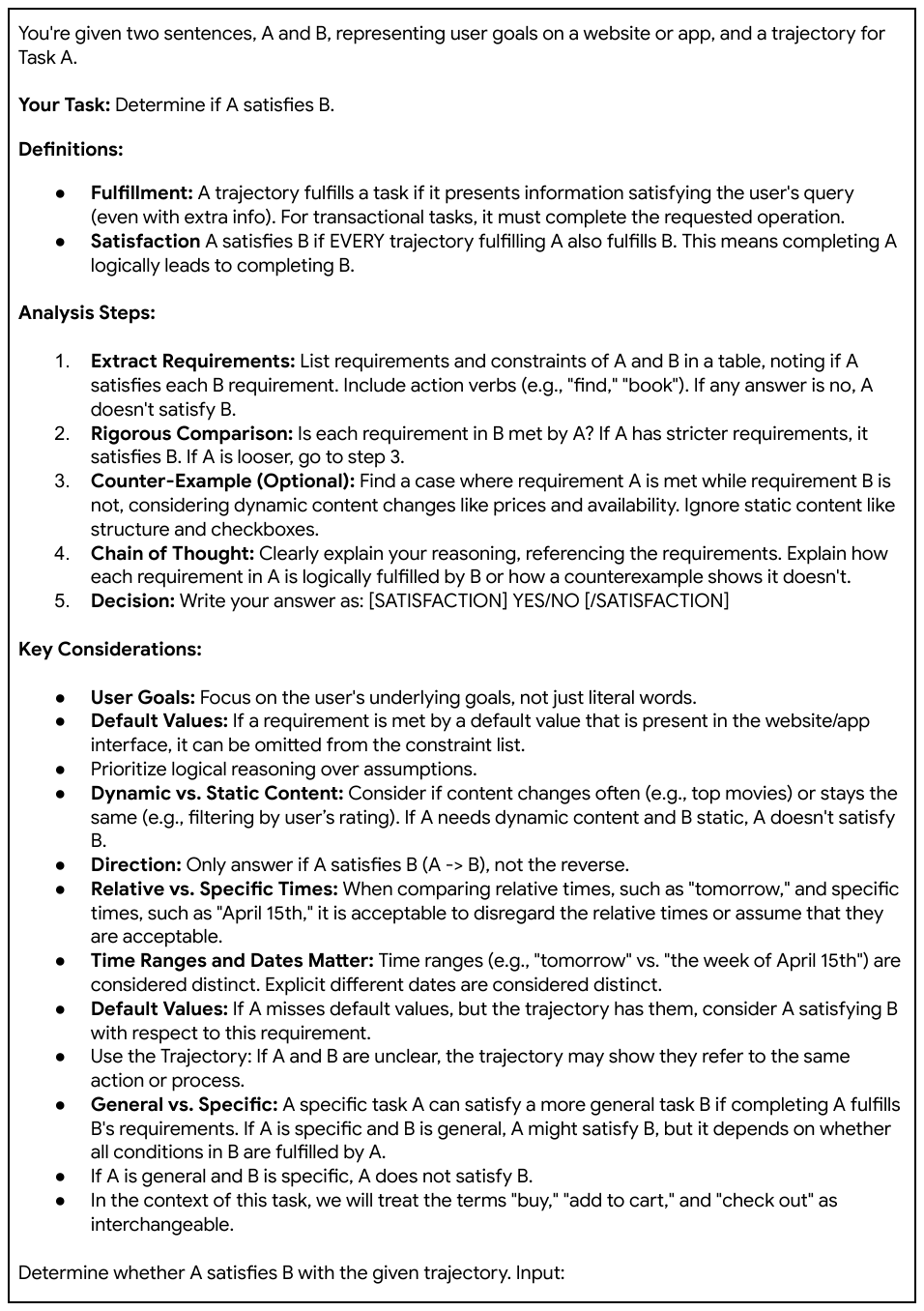}
\caption{Model instructions for evaluating Satisfaction relation between two task descriptions (A and B), given a corresponding trajectory. Includes web/mobile data format instructions and few-shot examples for experimentation (detailed on the next page).}
\Description{The figure contains a table with text displaying the instructions used to guide the GPT-4 model in automatically rating the satisfaction relation between two task descriptions.}
\label{fig:autorater_prompt}
\end{figure*}

\begin{figure*}[t]
\centering
\includegraphics[trim=0pt 90pt 0pt 70pt, clip, width=\textwidth]{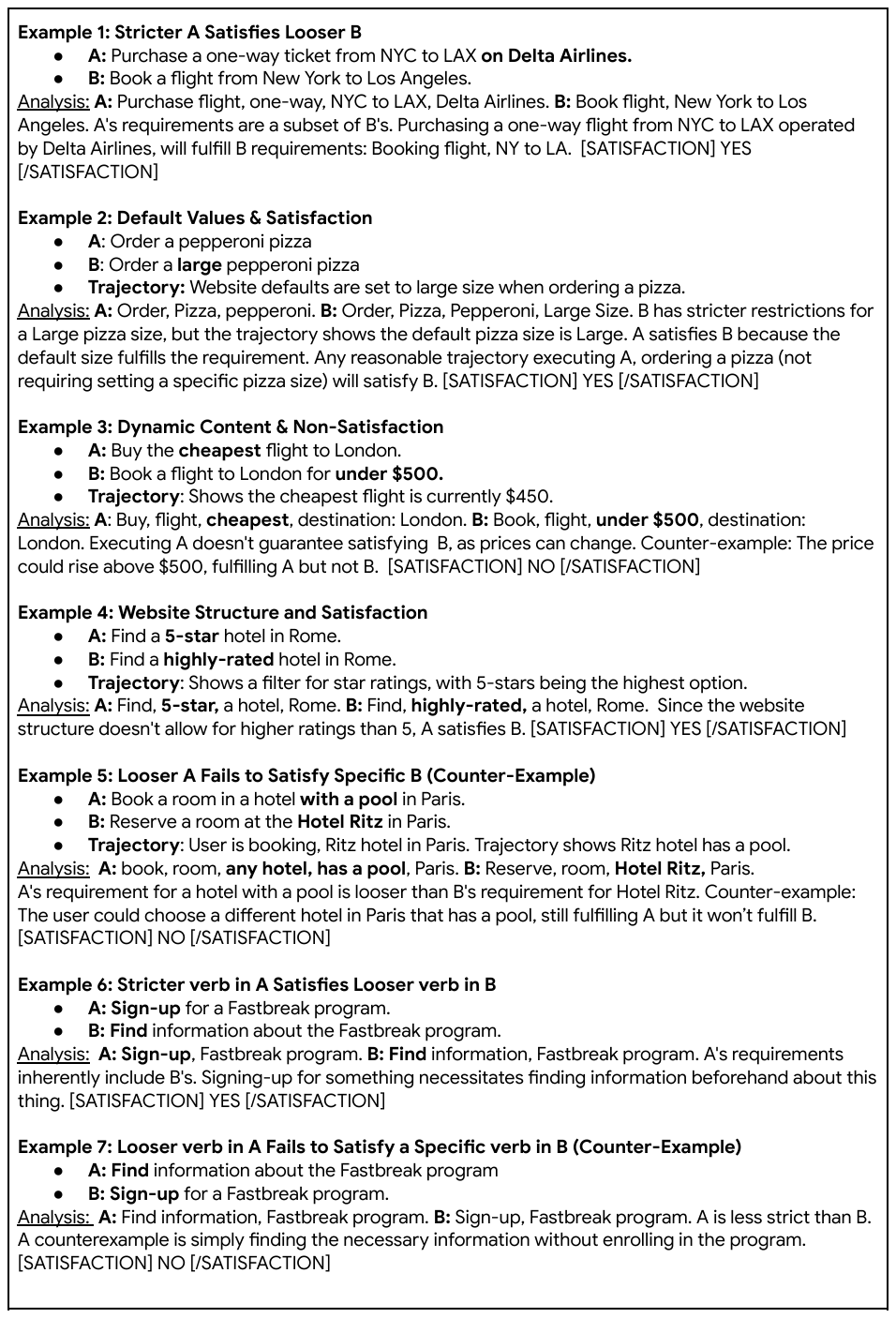}
\caption{Few-Shot exemplars demonstrating the Satisfaction Relation Prompt (Figure \ref{fig:autorater_prompt}). These simplified examples highlight the nuances of task satisfaction in real-world scenarios.}
\Description{The figure contains a table with text, displaying few-shot examples used in the prompt to guide the GPT-4 model in automatically rating two task descriptions.}
\label{fig:autorater_fewshot_examples}
\end{figure*}









\end{document}